\documentclass{article} \usepackage{iclr2027_conference,times}

\usepackage{amsmath,amsfonts,bm}

\def\eqref#1{equation~\ref{#1}}

\def\1{\bm{1}}

\DeclareMathAlphabet{\mathsfit}{\encodingdefault}{\sfdefault}{m}{sl}
\SetMathAlphabet{\mathsfit}{bold}{\encodingdefault}{\sfdefault}{bx}{n}

 \par

 \par

\usepackage{hyperref}
\usepackage{url}
\usepackage{graphicx}
\usepackage{xcolor}
\usepackage{booktabs}
\usepackage{array}
\usepackage{tabularx}
\usepackage{wrapfig}
\usepackage{tikz}
\usepackage{authblk}

\definecolor{prdcodebg}{HTML}{F6F8F7}
\definecolor{prdcodehead}{HTML}{202923}
\definecolor{prdcodeborder}{HTML}{D8DFDA}
\definecolor{prdcodeink}{HTML}{26312A}
\definecolor{prdcodemuted}{HTML}{6F7A73}
\definecolor{prdgreen}{HTML}{2D9144}
\definecolor{prdorange}{HTML}{D96D28}

\newcommand{\prdpriority}[2]{\begingroup
  \setlength{\fboxsep}{3pt}\colorbox{#1}{\textcolor{white}{\sffamily\bfseries\scriptsize #2}}\endgroup}

\newcommand{\prdsectiontitle}[2]{\par\noindent
  \begingroup
  \setlength{\fboxsep}{2.5pt}\colorbox{\prdaccent!13!white}{\textcolor{\prdaccent!78!black}{\sffamily\bfseries\scriptsize #1}}\endgroup
  \hspace{0.45em}{\ttfamily\bfseries\small #2}\par
  \vspace{3pt}}

\newcommand{\prdcodekv}[2]{\noindent
  \begin{tabularx}{\linewidth}{@{}>{\raggedright\arraybackslash}p{0.32\linewidth}@{\hspace{0.5em}}>{\raggedright\arraybackslash}X@{}}
  \textcolor{\prdaccent!80!black}{\ttfamily\bfseries\scriptsize #1:} &
  {\sffamily\footnotesize\color{prdcodeink}#2}
  \end{tabularx}\vspace{2pt}}

\newcommand{\prdcodedivider}{\vspace{4pt}{\color{prdcodeborder}\hrule height 0.45pt}\vspace{6pt}}

\newsavebox{\prdcodebox}
\newenvironment{prdcodeblock}[4]{\par\addvspace{7pt}\begingroup
  \def\prdaccent{#1}\begin{lrbox}{\prdcodebox}\begin{minipage}{\dimexpr\linewidth-1pt\relax}
    \begingroup
    \setlength{\fboxsep}{5pt}\noindent\colorbox{taskblue!55!black}{\parbox{\dimexpr\linewidth-2\fboxsep\relax}{\begin{tabularx}{\linewidth}{@{}Xr@{}}
      {\ttfamily\bfseries\small\textcolor{white}{#2}} &
      \prdpriority{taskblue}{#3}\hspace{0.65em}{\sffamily\scriptsize\textcolor{white!78}{#4}}
      \end{tabularx}
      }}\endgroup
    \par\vspace{7pt}
    \hspace*{5pt}\begin{minipage}{\dimexpr\linewidth-10pt\relax}
}{\end{minipage}\par\vspace{7pt}
  \end{minipage}\end{lrbox}\noindent
  \begin{tikzpicture}
  \node[
    draw=taskblue!55!black,
    fill=taskblue!3,
    line width=0.5pt,
    rounded corners=2pt,
    inner sep=0pt,
    outer sep=0pt
  ] {\usebox{\prdcodebox}};
  \end{tikzpicture}\par\endgroup\addvspace{7pt}}

\definecolor{taskblue}{RGB}{8,97,215}
\definecolor{darkgreen}{RGB}{0,110,60}
\definecolor{darkred}{RGB}{170,25,25}

\newsavebox{\findingboxbody}
\newenvironment{findingbox}[1]{\par\addvspace{4pt}\begin{lrbox}{\findingboxbody}\begin{minipage}{\dimexpr\linewidth-1pt\relax}\begingroup
  \setlength{\fboxsep}{5pt}\noindent\colorbox{taskblue!55!black}{\parbox{\dimexpr\linewidth-2\fboxsep\relax}{\color{white}\bfseries #1}}\endgroup
  \par\vspace{4pt}\hspace*{5pt}\begin{minipage}{\dimexpr\linewidth-10pt\relax}}{\end{minipage}\vspace*{4pt}\end{minipage}\end{lrbox}\noindent
  \begin{tikzpicture}
  \node[
    draw=taskblue!55!black,
    fill=taskblue!3,
    line width=0.5pt,
    rounded corners=2pt,
    inner sep=0pt,
    outer sep=0pt
  ] {\usebox{\findingboxbody}};
  \end{tikzpicture}\par\addvspace{4pt}}

\newcommand{\phoneworldrepo}{\url{https://github.com/EthanLeo-LYX/PhoneWorld}}
\newcommand{\phoneworldapks}{\url{https://huggingface.co/datasets/EthanLeoLYX/PhoneWorld-APKs}}

\newcommand{\huggingface}{\raisebox{-1.5pt}{\includegraphics[height=1.05em]{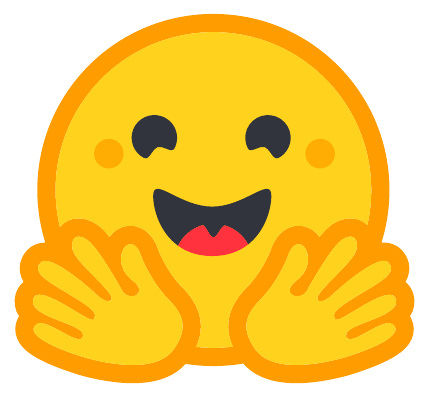}}}
\newcommand{\github}{\raisebox{-1.5pt}{\includegraphics[height=1.05em]{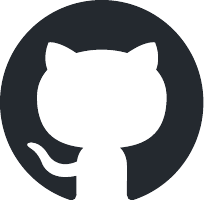}}}

\title{\raisebox{-4pt}{\includegraphics[height=1.8em]{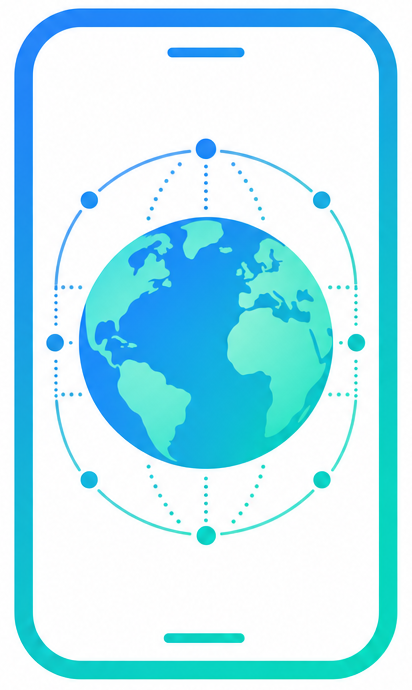}}~PhoneWorld: From Real-App Trajectories to Dynamic and Verifiable Environments for Phone-Use Agents}

\author[1,2]{\bf Yuxuan Liu{$^*$}}
\author[2]{\bf Xin Lai{$^*$}}
\author[2]{\bf Junyi Li{$^*$}}
\author[2]{\bf Pengyuan Lyu{$^*$}}
\author[2]{\bf Chenxin Li{$^*$}}

\author[2]{\bf Yiduo Guo}
\author[2]{\bf Zhengyao Fang}
\author[2]{\bf Yang Ding}
\author[2]{\bf Yi Zhang}
\author[2]{\bf Weinong Wang}
\author[2]{\bf Huawen Shen}
\author[2]{\bf Xingran Zhou}
\author[2]{\bf Liang Wu}
\author[2]{\bf Fei Tang}
\author[2]{\bf Sunqi Fan}
\author[2]{\bf Shangpin Peng}
\author[2]{\bf Zheng Ruan}
\author[2]{\bf Anran Zhang}

\author[2]{\bf Chengquan Zhang{$^\dag$}}
\author[2]{\bf Han Hu}

\author[4]{\bf Benyou Wang}
\author[1]{\bf Ji-Rong Wen}
\author[3]{\bf Rui Yan{$^\ddag$}}
\author[2,4]{\bf Zhengyang Tang{$^\ddag$}}

\affil[1]{Gaoling School of Artificial Intelligence, Renmin University of China}
\affil[2]{Tencent Hunyuan}
\makeatletter
\def\AB@affilsep{\protect\hspace{1.5em}\protect\Affilfont}
\makeatother
\affil[3]{Wuhan University}
\affil[4]{The Chinese University of Hong Kong, Shenzhen}

\newcommand\nnfootnote[1]{\begin{NoHyper}
  \renewcommand\thefootnote{}\footnote{#1}\addtocounter{footnote}{-1}\end{NoHyper}
}

\newcommand{\bulletitem}{\par\noindent\textbullet\hspace{0.5em}}

\iclrfinalcopy

\fancypagestyle{firstpage}{\fancyhead{}\lhead{\includegraphics[height=17pt]{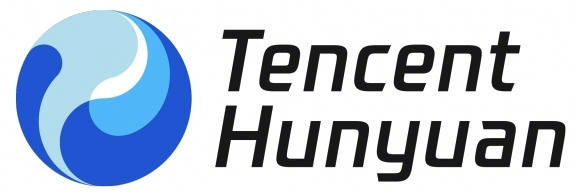}}\rhead{\textcolor[HTML]{4D4D4D}{\today}}\renewcommand{\headrulewidth}{1pt}\renewcommand{\headrule}{{\color[HTML]{4D4D4D}\hrule height \headrulewidth width \headwidth \vskip-\headrulewidth}}}

\begin{document}
\maketitle
\thispagestyle{firstpage}

\nnfootnote{$*$: Equal contribution.}
\nnfootnote{$\dag$: Project Lead.}
\nnfootnote{$\ddag$: Corresponding authors.}

\vspace{-0.4in}
\centerline{\href{https://github.com/EthanLeo-LYX/PhoneWorld}{\github~\texttt{Code}}\hspace{1.5em}
\href{https://huggingface.co/datasets/EthanLeoLYX/PhoneWorld-APKs}{\huggingface~\texttt{APKs}}}
\vspace{0.1in}

\begin{abstract}
Real applications provide the training setting closest to phone-agent deployment, but are
difficult to reset, scale safely, and verify programmatically. Static screenshots and interaction
trajectories preserve realistic evidence but cannot generate new experience. We introduce
\textbf{PhoneWorld}, a trace-grounded framework that converts such evidence into runnable,
resettable, and verifiable Android environments. PhoneWorld induces a usage-weighted interaction
skeleton from observed pages, transitions, and state-changing operations; translates it into a
behavior-grounded app specification; realizes the specification through an autonomous
build--inspect--repair loop; and synthesizes executable tasks
with programmatic verifiers. The resulting suite spans 34 consumer-facing apps across 16 domains
and supports an audited online benchmark, verified trajectory generation, and online RL through
common reset and verification interfaces. Evaluations with diverse general and open-source GUI
agents show that PhoneWorld supports reliable end-to-end online interaction and exposes
capabilities complementary to AndroidWorld. Controlled SFT experiments further show that
PhoneWorld trajectories complement AndroidWorld supervision, transfer across online and offline
benchmarks, and become more effective as data volume and app coverage increase. Under a matched
RL budget, combining PhoneWorld mock-app rollouts with real-app rollouts improves performance over
real-app RL alone on both real-phone tasks and AndroidWorld. Together, these results demonstrate
that trace-grounded executable abstraction can bridge realistic mobile behavior and scalable
agent learning, turning limited real-app evidence into a growing supply of controllable and
verifiable environments for training and evaluation.

\end{abstract}

\section{Introduction}

Mobile agents have advanced rapidly from single-screen perception---such as screen question
answering, UI-element grounding, and isolated clicking---to autonomous systems that can execute
complex, multi-step tasks with little human intervention
~\citep{wang2024mobile, wang2025mobile, zhang2025appagent, qin2025ui, wang2025ui,
zhou2025mai, xu2026mobilev3, liu2025mobilesteward}. Learning these capabilities depends critically on interaction data,
whose scale and diversity are determined by the environments in which agents act. Because
deployment ultimately occurs in real applications, improving real-world assistance requires
agents to learn authentic interfaces, content, navigation structures, and state transitions;
direct real-app interaction therefore provides the closest training setting to deployment.
Alongside this shift in capability, mobile-agent research has expanded from single-screen
understanding to learning from static offline trajectories, and more recently to dynamic data
collection and online training in executable mobile environments~\citep{xu2025mobilerl,
shi2025mobilegui, luo2025gui, chen2025step,
xi2026toolgym}.

The shift from static supervision to dynamic interaction, however, exposes a fundamental tension:
the environments closest to real-world deployment are also the most difficult to use for
training at scale. Concretely, production apps are difficult to reset and run safely in parallel,
and rarely expose programmatic task verification, limiting scalable data collection and online
training. AndroidWorld~\citep{rawles2025androidworld} and
MobileWorld~\citep{kong2025mobileworld} enable reproducible online evaluation through reset and
programmatic verification, but rely primarily on relatively simple open-source or system apps,
leaving a substantial gap from widely deployed consumer apps in functional scope, page structure,
and interaction logic. AndroidAgentArena ~\citep{chai2025a3} and
AndroidDaily~\citep{sui2026androiddaily} instead evaluate live production or closed-source apps,
narrowing the realism gap. Yet such environments remain difficult to reset and do not readily
support programmatic verification.
This leaves three questions:
(1) \textit{Static-to-Dynamic Environment Synthesis}: Can static real-app screenshots and
trajectories be transformed into runnable mock apps that preserve their task-relevant pages,
navigation paths, operations, and state changes;
(2) \textit{Controllable and Verifiable Environment Support}: Can these apps be equipped with
resettable state, executable tasks, and programmatic verifiers to support evaluation, verified
data collection, and online RL; and
(3) \textit{Transferable Training Utility}: Can supervision and rewards produced in these
environments transfer beyond the mock apps and complement real-app-derived training?

\begin{figure*}[t]
\centering
\includegraphics[width=\linewidth]{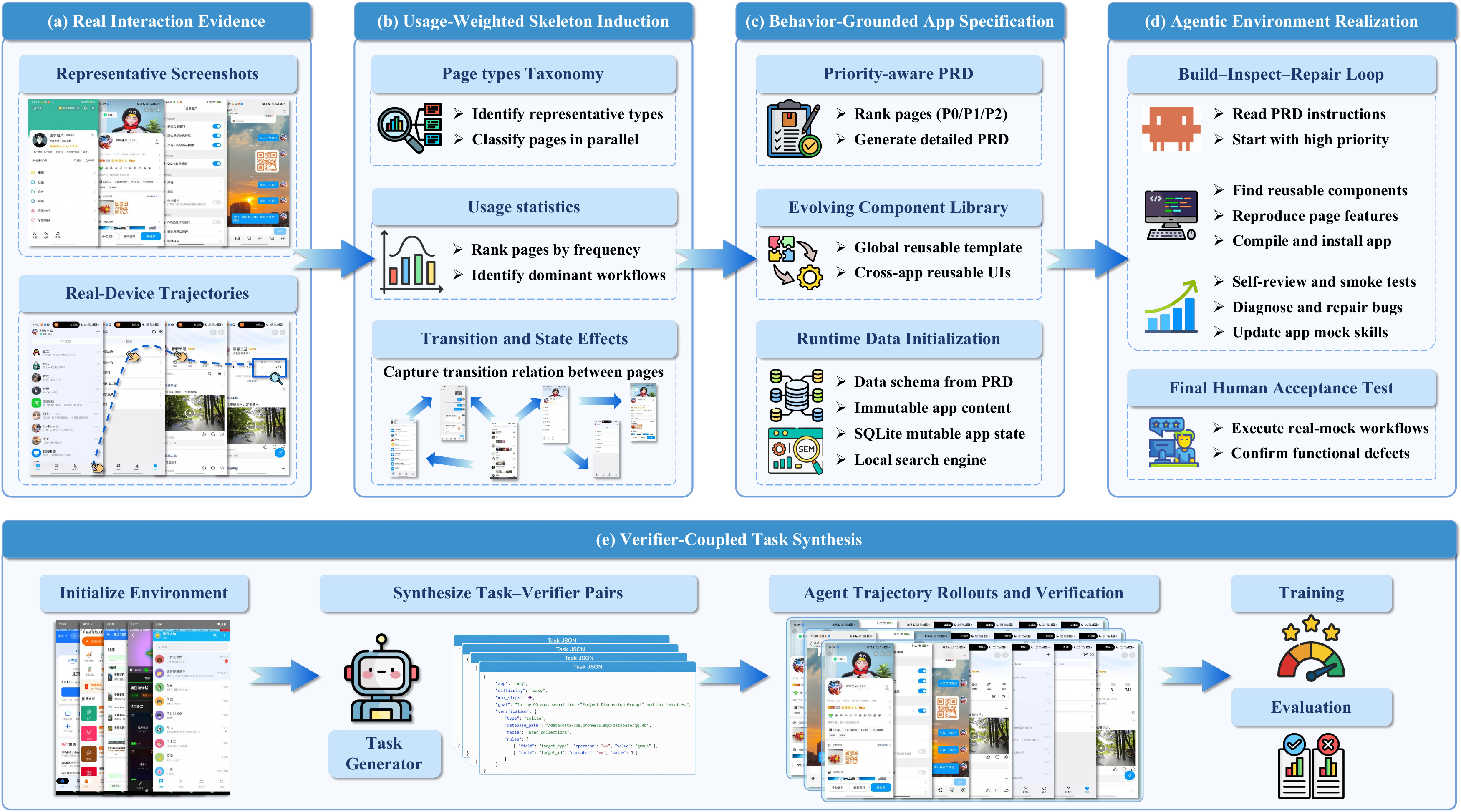}
\caption{\textbf{PhoneWorld transforms static real-app evidence into dynamic online environments.}
\textbf{(a) Real Interaction Evidence}: representative screenshots and exploratory-use trajectories collected from a real app.
\textbf{(b) Usage-Weighted Skeleton Induction}: page types, visitation frequencies, navigation edges, and state-changing operations are induced from the observations.
\textbf{(c) Behavior-Grounded App Specification}: the skeleton is translated into prioritized page, transition, content, and state-effect contracts together with reusable interaction primitives.
\textbf{(d) Agentic Environment Realization}: a coding agent implements and iteratively repairs a runnable Android environment, followed by final human acceptance testing.
\textbf{(e) Verifier-Coupled Task Synthesis}: tasks and programmatic verifiers are synthesized from the same content and mutable state, enabling online evaluation, rollout generation, and RL.}
\label{fig:teaser}
\end{figure*}

To bridge this realism--controllability gap, we introduce \textbf{PhoneWorld}, a trace-grounded environment
synthesis framework that distills the task-relevant structure and behavior of real applications
into dynamic, resettable, and programmatically verifiable mobile environments. As shown in Figure~\ref{fig:teaser},
PhoneWorld operationalizes this framework through four stages:
(1) \textbf{Usage-Weighted Skeleton Induction} identifies behaviorally meaningful
page types and recovers their visitation frequencies, navigation graph, and state-changing
operations.
(2) \textbf{Behavior-Grounded App Specification} translates this skeleton into prioritized page
PRDs, reusable functional components, and runtime content and state specifications.
(3) \textbf{Agentic Environment Realization} autonomously implements and repairs the
specification as a runnable Android app, followed by final human acceptance testing.
(4) \textbf{Verifier-Coupled Task Synthesis} derives executable tasks and programmatic verifiers
from the same content and mutable state, enabling verified rollouts for data generation,
reinforcement learning, and evaluation.

The resulting PhoneWorld suite comprises 34 consumer-facing Android environments spanning 16
domains. Each environment couples a runnable interface with deterministic content, resettable
mutable state, grounded tasks, and programmatic verifiers, enabling the same infrastructure to
support evaluation, verified rollout generation, and online RL. The current suite contains a
120-task audited benchmark, a pool of 7,936 executable tasks, and 3,354 verifier-confirmed
successful trajectories. Crucially, these environments provide transferable training benefits. Under
a 36K-step data budget, replacing 10K steps of AndroidWorld supervision with 10K
PhoneWorld rollout steps improves success by 52.5 points on PhoneWorld and 14.7 points on
AndroidWorld. Holding the PhoneWorld budget at 10K steps, expanding app coverage from 5 to 34
yields further gains of 18.3 points on PhoneWorld and 10.4 points on AndroidWorld, demonstrating
that environment diversity contributes beyond data volume. This benefit also extends to online
learning: under the same 50-step RL budget, a 50/50 real/mock rollout mixture improves a
separate 4B agent over real-app RL alone, from 40.67\% to 45.33\% on a 150-task real-phone
evaluation and from 77.2\% to 83.2\% on AndroidWorld. Finally, a four-app audit shows that the
generated apps preserve 97.7\% rendered, 92.4\% navigable, and 73.0\% functional coverage of
the observed page-visit mass.

Our contributions are threefold:

\bulletitem We introduce a \textbf{trace-grounded executable abstraction framework} that transforms static real-app screenshots and trajectories into dynamic, stateful Android environments through Usage-Weighted Skeleton Induction, Behavior-Grounded App Specification, and autonomous Agentic Environment Realization with final human acceptance testing.

\bulletitem We construct \textbf{PhoneWorld}, a unified suite of 34 consumer-facing Android environments across 16 domains. Unified reset, task, and verifier interfaces jointly support \textbf{verified data generation, online reinforcement learning, and task evaluation}, including a 120-task audited benchmark.

\bulletitem We demonstrate that \textbf{PhoneWorld provides effective and transferable training signals}. Controlled supervised-training experiments show gains on both PhoneWorld and AndroidWorld under a matched budget, with further improvements from broader app coverage at a fixed data budget. Under the same total RL budget, incorporating PhoneWorld mock-app interaction into a mixed real--mock recipe further improves both real-phone evaluation and AndroidWorld over real-app RL alone.

\section{Related Work}
Training capable phone-use agents requires more than static screenshots or demonstration
trajectories: agents need a large supply of interactive environments in which they can execute
multi-step actions, observe state changes, and receive reliable automatic
feedback. This requirement has motivated a shift from static mobile benchmarks and offline
trajectory learning~\citep{deng2024mobile, xu2025mobile, xu2025androidlab, zhang2026tongui,
lu2025videoagenttrek, wang2026opencua, huang2025mobileipl, liu2026come} toward online interaction and reinforcement learning
for GUI agents~\citep{xu2025mobilerl, shi2025mobilegui, wang2025ui, gu2025uivenus,
venusteam2026uivenus15, xu2026appdeltaworld, tang2026training}. The central artifact in this setting is therefore no longer only a
task set or a dataset, but a verifiable environment that can repeatedly produce new interaction
experience.

Existing mobile-agent benchmarks have made important progress toward verifiable online
environments. AndroidWorld~\citep{rawles2025androidworld} and MobileWorld
~\citep{kong2025mobileworld} provide resettable Android tasks with programmatic checks, while
MobileBench-OL~\citep{wu2026mobilebench} emphasizes practical mobile task execution. Their
controllability, however, typically depends on a limited set of simple open-source, system, or
locally instrumented apps. Conversely, Android Agent Arena~\citep{chai2025a3} and
AndroidDaily~\citep{sui2026androiddaily} move closer to live or closed-source production apps,
improving realism but making robust reset, safe parallel execution, and deterministic
verification harder. This exposes a core realism--controllability trade-off: real apps are most
relevant for training but hardest to scale.

A complementary line of work studies how to generate or synthesize interactive environments for
agents at scale~\citep{zala2024envgen, wang2026agent, cao2026gui, xu2026mobile}. InfiniteWeb
~\citep{zhang2026infiniteweb} generates functional multi-page websites with task-centric
specifications and verifiable evaluators, AutoWebWorld~\citep{wu2026autowebworld} translates
finite-state environment models into interactive websites, GUI Exploration Lab
~\citep{yan2026gui} builds controllable simulators for studying screen navigation, and
Agent-World~\citep{dong2026agentworld} synthesizes tool- and database-based environments for
general agent training and evaluation. PhoneWorld brings this idea to phone-use agents
by grounding environment synthesis in real mobile interaction evidence: it abstracts observed
pages, navigation paths, operations, and state effects from real-app trajectories, then realizes
them as executable Android environments that support evaluation, verified trajectory generation,
and online RL through the same reset and verification interfaces.

\section{Method: Trace-Grounded Executable Abstraction}
\label{sec:method}
PhoneWorld frames environment construction as \emph{trace-grounded executable abstraction}: from
partial observations of a production app, it preserves task-relevant pages, navigation, visible
semantics, and state effects while adding reset and verification interfaces. Rather than cloning
the full service, PhoneWorld induces a usage-weighted interaction skeleton, specifies executable
behavior, realizes it as a stateful Android environment, and synthesizes tasks and verifiers from
the resulting state space. Figure~\ref{fig:teaser} summarizes the four stages, and
Figure~\ref{fig:qq_case} gives a concrete example.

\subsection{From Static Traces to Dynamic Environments}
For a target app, let \(\mathcal{S}\) denote its representative screenshot corpus and let
\(\mathcal{T}=\{\tau_i\}_{i=1}^{N}\) denote static real-device trajectories. A trajectory is
\(\tau_i=(g_i,o_{i,0},a_{i,0},o_{i,1},\ldots,a_{i,L_i-1},o_{i,L_i})\), where \(g_i\) is the user
instruction, \(o_{i,t}\) is a screenshot, and \(a_{i,t}\) is a touch or system action. Together,
\(\mathcal{D}=(\mathcal{S},\mathcal{T})\) provides only a static view of the
realistic production app. It captures rendered states and observed transitions, but does not
constitute an executable model of the app's interaction dynamics---how it responds to new action
sequences, evolves persistent state, or supports repeated exploration.

PhoneWorld first induces a \textbf{Usage-Weighted Interaction Skeleton}
\(\mathcal{K}=(\mathcal{P},\mathcal{G},\mathcal{Z})\), where
\(\mathcal{P}\) is a page-type inventory, \(\mathcal{G}\) is a weighted navigation graph, and
\(\mathcal{Z}\) captures action-conditioned state effects. \textbf{Behavior-Grounded App Specification} then translates
this skeleton into an executable app specification \(\boldsymbol{\Sigma}\), which defines the page-level
interface, navigation logic, reusable interaction primitives, content organization, mutable-state
semantics, and reset behavior required by the selected workflows. \textbf{Agentic Environment
Realization} instantiates this specification as a runnable and resettable surrogate environment
\(\widehat{\mathcal{E}}\). Finally, \textbf{Verifier-Coupled Task Synthesis} derives executable tasks,
programmatic verifiers, and verified rollouts from the resulting environment:
\begin{equation}
\mathcal{D}
\xrightarrow{\;\mathrm{induce}\;}
\mathcal{K}
\xrightarrow{\;\mathrm{specify}\;}
\boldsymbol{\Sigma}
\xrightarrow{\;\mathrm{realize}\;}
\widehat{\mathcal{E}}
\xrightarrow{\;\mathrm{synthesize}\;}
\bigl(\mathcal{Q},\mathcal{V},\mathcal{R}^{+}\bigr).
\label{eq:pipeline_formal}
\end{equation}
Here, \(\mathcal{Q}\) contains executable tasks, \(\mathcal{V}\) their programmatic verifiers,
and \(\mathcal{R}^{+}\) verifier-confirmed successful trajectories. The distinction between
\(\mathcal{K}\) and \(\boldsymbol{\Sigma}\) is central: \(\mathcal{K}\) determines what task-relevant
behavior should be preserved from the observed evidence, whereas \(\boldsymbol{\Sigma}\) specifies how
that behavior should be implemented as an executable app. We refer to
\(\widehat{\mathcal{E}}\) as a behavioral surrogate because it preserves the task-relevant
interaction surface of the production app for controlled learning, rather than attempting to
reproduce the complete product.

\subsection{Usage-Weighted Skeleton Induction}
The induction stage determines \emph{what should survive abstraction}. Screenshots reveal the
visual organization and available affordances of individual pages, while trajectories reveal the
app's operational structure: how frequently pages are visited, how they are connected within
common workflows, and how user actions change the app state. PhoneWorld combines
these signals into an \emph{interaction skeleton} that prioritizes empirically important behavior
instead of treating every production feature as equally necessary.

We first organize representative screenshots into an extensible, app-specific taxonomy of
approximately 25--30 page types, such as profile, search, and detail pages. A vision-language
model (VLM) then maps the screenshot corpus into this taxonomy. The taxonomy abstracts transient
visual variations into behaviorally meaningful page types. Screens showing different content
instances or scroll positions are grouped together when they support the same task-relevant
interactions, while visually similar screens remain separate when their available operations or
navigation roles differ.

Formally, let \(c(o)\in\mathcal{P}\) map a screenshot to a page type in the recovered
taxonomy. We estimate normalized page frequency and transition weight as
\begin{align}
\widehat f(p)
&=
\frac{1}{M}
\sum_{i=1}^{N}\sum_{t=0}^{L_i}
\mathbf{1}\!\left[c(o_{i,t})=p\right],
\qquad
M=\sum_{i=1}^{N}(L_i+1),
\label{eq:page_frequency}\\
w(p,q)
&=
\sum_{i=1}^{N}\sum_{t=0}^{L_i-1}
\mathbf{1}\!\left[c(o_{i,t})=p,\;c(o_{i,t+1})=q\right].
\label{eq:transition_weight}
\end{align}
These statistics define a weighted page graph
\(\mathcal{G}=(\mathcal{P},\mathcal{E}^{\mathrm{nav}},w)\), with
\((p,q)\in\mathcal{E}^{\mathrm{nav}}\) whenever \(w(p,q)>0\). The node frequencies quantify
how prominently each page appears in observed use, while the edge weights capture how often
page pairs are traversed consecutively. Together, they recover the dominant pages and navigation
paths of the app.

Page visitation and navigation statistics alone do not capture the semantic effects of operations
such as favorite, follow or order placement. We therefore
associate important interactions with state-effect contracts \(z=(p,a,\psi_z)\in\mathcal{Z}\),
where \(p\) is the source page, \(a\) is the user action, and \(\psi_z\) describes the semantic
post-condition to preserve.
For example, a favorite action should make the target entity appear in the user's collection, and
an add-to-cart action should update both cart membership and quantity. These contracts connect
the observed interaction skeleton to the mutable-state design used later for verification.
A worked construction example appears in Appendix~\ref{app:worked_example}.

\subsection{Behavior-Grounded App Specification}
The interaction skeleton identifies what to preserve but not how to implement it. PhoneWorld
therefore converts it into an executable app specification \(\boldsymbol{\Sigma}\) covering three
aspects:\emph{ how to design app, how to implement functional components, and how to prepare runtime data.}

We first map the page-frequency ranking to three construction tiers: P0 pages and their connecting
paths form the behavioral backbone, P1 pages extend common workflows, and P2 pages cover selected
long-tail functionality. For each selected page, a VLM turns representative screenshots into a
structured PRD specifying its layout, controls, transitions, content fields, and state effects.
Two condensed examples of the resulting page specifications are provided in
Appendix~\ref{app:prd_examples}.
Recurring requirements are mapped to an evolving library of reusable UI and logic components,
such as search, feeds, comments, messaging, carts, and checkout. As more apps are constructed,
recurring patterns and validated fixes are consolidated into this library, allowing subsequent
apps to inherit increasingly reliable components while reducing repeated implementation effort.
Finally, we prepare runtime data by separating immutable app content \(C\) from mutable user
state \(m_t\). For each app, its PRDs determine the data schema and fields required by the
supported pages and workflows. The immutable store contains task-relevant entities, attributes,
and relations for browsing, detail-page rendering, and deterministic offline BM25 search without
external services. The SQLite-backed mutable store records action-induced updates, such as
favorites, comments, messages, cart contents, and orders, and exposes the same state to both the
interface and downstream verifiers. Before each task, reset restores a fixed initial snapshot,
providing reproducible initial conditions and repeatable execution. Together, these elements
preserve task-relevant app design and behavior without reproducing the entire product.

\subsection{Agentic Environment Realization}
A coding agent translates the app specification into Kotlin and Jetpack Compose through an
iterative build--inspect--repair process. It first scaffolds the Android project and data layer,
instantiates applicable components from the reusable library, and implements the P0 pages and
their end-to-end paths before extending coverage to P1 and selected P2 functionality. The agent
then connects page routes, interface controls, content queries, state updates, and reset logic
according to the PRDs.

After each implementation pass, the agent compiles and installs the APK and runs smoke tests over
launch, navigation, search, and representative state-changing operations. Compiler messages,
runtime failures, broken routes, inactive controls, rendering problems, and inconsistent state
updates are detected by the agent and used to guide the next repair pass. This closed loop allows
the agent to diagnose failures, revise the implementation, and iteratively optimize the app until
the specified workflows execute correctly, without human intervention during construction.
Recurring failures and validated solutions are incorporated into the construction checklist and
reusable component library, so later apps benefit from earlier implementation experience. The
resulting app provides deterministic local content, persistent action effects, and a reset
operation that restores a known initial snapshot, allowing the same workflows to be executed
repeatedly. Finally, human acceptance testing checks whether any functional defects remain in the
principal real--mock workflows.

\subsection{Verifier-Coupled Task Synthesis}
A runnable surrogate becomes a learning environment only when goals, executable paths, and
success conditions agree. PhoneWorld therefore synthesizes each task jointly with its verifier
rather than generating an instruction first and attaching an evaluator afterward. Immutable
content determines which entities and answers exist, page specifications constrain the available
paths and controls, and the mutable-state schema exposes outcomes that can be checked.

A task \(q=(g_q,x_q^0,H_q,v_q)\) consists of a natural-language goal \(g_q\), a reset
configuration \(x_q^0\), a maximum interaction horizon \(H_q\), and a programmatic verifier
\(v_q\). Given \(q\), an agent policy \(\pi\) interacts with the environment
\(\widehat{\mathcal{E}}\) for at most \(H_q\) steps and produces a trajectory
\(\xi=(o_0,a_0,\ldots,o_T,y_T,m_T)\), where \(o_t\) and \(a_t\) are the screen observation and
action at step \(t\), \(y_T\) is an optional final answer, and \(m_T\) is the final mutable app
state. The verifier maps this trajectory to a binary task outcome:
\begin{equation}
(g_q,x_q^0)
\xrightarrow[\widehat{\mathcal{E}},\,T\leq H_q]{\;\pi\;}
\xi
\xrightarrow{\;v_q\;}
V_q(\xi)=v_q(y_T,m_T)\in\{0,1\}.
\label{eq:task_environment_closure}
\end{equation}
This joint construction forms a task--environment--verifier closure: every requested entity is
grounded in the app content, every required operation is supported by the app specification, and
every success condition is observable in the final answer or environment state. Answer verifiers
test grounded values in \(y_T\), while database verifiers require \(m_T\) to contain records
satisfying task-specific predicates. The resulting binary outcome supports evaluation, filtering
successful trajectories for supervised learning, and terminal rewards for online RL.
Appendix~\ref{app:verifier_details} provides the complete verifier semantics and construction
procedure.

\subsection{PhoneWorld Environment Suite}
Applying this pipeline and the evolving reusable-component library, we construct PhoneWorld as a
unified suite of 34 mock Android apps spanning 16 consumer-facing domains. Rather than serving as
standalone APKs, all apps follow a common environment interface that couples deterministic
content with resettable mutable state, executable task definitions, and programmatic verifiers.
This common interface preserves app-specific pages, functions, and data while allowing agents to
interact with every environment through the same task-execution and reset protocol.

Each environment supports three connected uses. First, tasks are synthesized from the app's
available content, operations, and state schema, so their completion conditions can be checked
programmatically. Second, agents can execute these tasks at scale, with verifier-confirmed
trajectories retained as training data. Third, reproducible initialization and programmatic
terminal rewards provide a stable environment for online RL. The current suite contains a
120-task audited benchmark, a pool of 7,936 generated tasks, and 3,354 verified successful
trajectories totaling 36,193 interaction steps, as summarized in
Table~\ref{tab:phoneworld_suite} in Appendix~\ref{app:experimental_details}.

\section{Experiments}
Our experiments follow three main questions.
\textbf{Q1}: can heterogeneous agents interact with and be evaluated by PhoneWorld?
\textbf{Q2}: do verified trajectories collected from PhoneWorld improve supervised training?
\textbf{Q3}: can the environments themselves provide useful online RL experience on top of real-app RL?

\subsection{PhoneWorld supports interactive and verifiable online evaluation}
We evaluate seven general agents and five open-source GUI agents on the complete
120-task PhoneWorld benchmark and 116-task AndroidWorld benchmark.
We train \textbf{PWAgent} by supervised fine-tuning Qwen3.5-9B on a 36K-step mixture of 26K
AndroidWorld and 10K PhoneWorld steps.
Both online evaluations report task success across easy, medium, and hard subsets; an attempt is
successful only when its programmatic verifier returns full credit.

\begin{table*}[t]
\centering
\setlength{\belowcaptionskip}{4pt}
\small
\setlength{\tabcolsep}{3.5pt}
\caption{\textbf{Primary online evaluation by task difficulty.} We report strict task success
rate (\%) on both PhoneWorld and AndroidWorld benchmarks. \emph{Steps} is the mean number of
dispatched device actions over the tasks an agent solved on that benchmark.}
\label{tab:main_result}
\begin{tabular}{l ccccc ccccc}
\toprule
\midrule
\textbf{Model / setting} &
\multicolumn{5}{c}{\textbf{PhoneWorld}} &
\multicolumn{5}{c}{\textbf{AndroidWorld}} \\
\cmidrule(lr){2-6}\cmidrule(lr){7-11}
& \textbf{Easy} & \textbf{Medium} & \textbf{Hard} & \textbf{Overall} & \textbf{Steps}
& \textbf{Easy} & \textbf{Medium} & \textbf{Hard} & \textbf{Overall} & \textbf{Steps} \\
\midrule
\multicolumn{11}{l}{\textit{General-agent references}} \\
Claude Opus 4.8 & 100.0 & 95.1 & 85.7 & 94.2 & 11.9 & 91.8 & 80.6 & 31.6 & 78.4 & 11.2 \\
Gemini 3.5 Flash & 97.7 & 87.8 & 91.4 & 92.5 & 12.1 & 95.1 & 80.6 & 63.2 & 85.3 & 10.4 \\
Kimi-K3 & 86.4 & 90.2 & 88.6 & 88.3 & 12.8 & 86.9 & 66.7 & 36.8 & 72.4 & 10.8 \\
GPT-5.5 & 84.1 & 82.9 & 91.4 & 85.8 & 12.8 & 91.8 & 66.7 & 31.6 & 74.1 & 12.8 \\
MiMo 2.5 & 84.1 & 82.9 & 82.9 & 83.3 & 13.2 & 82.0 & 41.7 & 21.1 & 59.5 & 11.7 \\
Seed 2.0 & 77.3 & 75.6 & 74.3 & 75.8 & 14.7 & 80.3 & 52.8 & 21.1 & 62.1 & 10.6 \\
DeepSeek-V4 Flash & 72.7 & 75.6 & 68.6 & 72.5 & 15.9 & 73.8 & 52.8 & 15.8 & 57.8 & 10.5 \\
\midrule
\multicolumn{11}{l}{\textit{Open-source GUI-agent references}} \\
GUI-Owl-1.5-8B & 56.8 & 58.5 & 51.4 & 55.8 & 13.2 & 72.1 & 52.8 & 21.1 & 57.8 & 10.0 \\
UI-Venus-1.5-8B & 63.6 & 63.4 & 40.0 & 56.7 & 10.1 & 65.6 & 33.3 & 15.8 & 47.4 & 7.4 \\
UI-Venus-Navi-7B & 56.8 & 56.1 & 34.3 & 50.0 & 15.6 & 62.3 & 30.6 & 15.8 & 44.8 & 9.3 \\
UI-TARS-7B-SFT & 50.0 & 53.7 & 31.4 & 45.8 & 10.8 & 37.7 & 13.9 & 5.3 & 25.0 & 7.9 \\
GUI-Owl-7B & 54.5 & 26.8 & 31.4 & 38.3 & 15.4 & 60.7 & 41.7 & 10.5 & 46.6 & 8.4 \\
\midrule
\multicolumn{11}{l}{\textit{Our trained agent}} \\
PWAgent & 77.3 & 63.4 & 51.4 & 65.0 & 11.4 & 83.6 & 66.7 & 42.1 & 71.6 & 10.9 \\
\midrule
\bottomrule
\end{tabular}
\end{table*}

Table~\ref{tab:main_result} demonstrates that PhoneWorld functions as an interactive and
programmatically verifiable online benchmark. All 13 agents can be evaluated end to end across
the 120 tasks, validating the integration of environment reset, multi-step interaction,
persistent state updates, and automatic verification. The broad score range---72.5--94.2\% for
general agents and 36.6--56.7\% for open-source agents---shows that the benchmark is discriminative.
All five open-source agents and PWAgent perform worse on hard than easy tasks; for example,
PWAgent decreases from 77.3\% to 51.4\%. PWAgent reaches 65.0\% on PhoneWorld and 71.6\% on
AndroidWorld, exceeding the strongest reported open-source reference by 8.3 and 13.8 points,
respectively.

The \emph{Steps} columns measure execution efficiency on solved tasks. For 12 of the 13 model
rows, successful PhoneWorld tasks require more actions than AndroidWorld tasks, consistent with
longer consumer-app workflows. Among general agents, stronger PhoneWorld performance also tends
to coincide with shorter successful trajectories: Claude Opus 4.8 averages 11.9 actions, compared
with 15.9 for DeepSeek-V4 Flash. PWAgent averages 11.4 actions, fewer than GUI-Owl-1.5-8B
(13.2) and UI-Venus-Navi-7B (15.6), so its gain does not come from taking longer successful
paths.

More importantly, relative model rankings change across the two benchmarks. Among general agents,
Claude Opus 4.8 leads Gemini 3.5 Flash on PhoneWorld (94.2\% vs.\ 92.5\%) but trails it on
AndroidWorld (78.4\% vs.\ 85.3\%). Among open-source agents, UI-Venus-1.5-8B slightly
outperforms GUI-Owl-1.5-8B on PhoneWorld (56.7\% vs.\ 55.8\%) but trails it on AndroidWorld
(47.4\% vs.\ 57.8\%); UI-Venus-Navi-7B similarly leads GUI-Owl-7B on PhoneWorld
(50.0\% vs.\ 36.6\%) but not on AndroidWorld (44.8\% vs.\ 46.6\%). PhoneWorld emphasizes
Chinese-language consumer apps, while AndroidWorld primarily covers English-language open-source
and system apps. These rank reversals indicate that the benchmarks expose different agent
strengths rather than differing only in overall difficulty.

\subsection{PhoneWorld-generated data improves supervised training}
We evaluate PhoneWorld-generated data through three controlled questions: whether it complements
AndroidWorld data, whether performance scales with the number of verified PhoneWorld steps, and
whether broader app coverage helps at a fixed PhoneWorld budget. PhoneWorld and AndroidWorld are
online benchmarks, whereas AndroidControl and Internal-MobileBench (IMB) are offline benchmarks.
Together, they cover Chinese- and English-language settings as well as both simple benchmark apps
and realistic consumer apps.

\vspace{-5pt}
\paragraph{Mixing PhoneWorld with AndroidWorld.}
\begin{wraptable}{r}{0.49\columnwidth}
\vspace{-20pt}
\centering
\setlength{\belowcaptionskip}{4pt}
\small
\setlength{\tabcolsep}{3.2pt}
\caption{\textbf{Matched-budget data composition.} All settings use 36,193 training steps. PW
and AW report online task success; IMB and AC report offline step success.}
\label{tab:matched_budget_composition}
\begin{tabular*}{\linewidth}{@{\hspace{4pt}\extracolsep{\fill}}lrrrr@{\hspace{4pt}}}
\toprule
\midrule
\textbf{Composition} & \textbf{PW} & \textbf{AW} & \textbf{IMB} & \textbf{AC} \\
\midrule
36K AW & 12.5 & 56.9 & 15.5 & 53.7 \\
26K AW + 10K PW & 65.0 & \textbf{71.6} & \textbf{33.2} & \textbf{59.7} \\
36K PW & \textbf{73.3} & 46.6 & \textbf{33.2} & 59.3 \\
\midrule
\bottomrule
\end{tabular*}
\end{wraptable}
Table~\ref{tab:matched_budget_composition} compares three mixtures under the same 36,193-step
total training budget: 36k AndroidWorld steps, 26k AndroidWorld + 10k PhoneWorld
steps, and 36k PhoneWorld steps.
Replacing 10K of the 36K AndroidWorld steps with PhoneWorld rollouts improves PhoneWorld by 52.5
points, AndroidWorld by 14.7 points, Internal-MobileBench by 17.7 points, and AndroidControl by 6.0
points, without increasing the total budget. Full replacement yields the best PhoneWorld score
(73.3\%) but lowers AndroidWorld to 46.6\%. The central result is therefore
\emph{complementarity}: PhoneWorld adds transferable interaction diversity, while retaining
real-app-derived supervision is important for the strongest joint performance.

\begin{figure*}[t]
\vspace{-15pt}
\centering
\includegraphics[width=\linewidth]{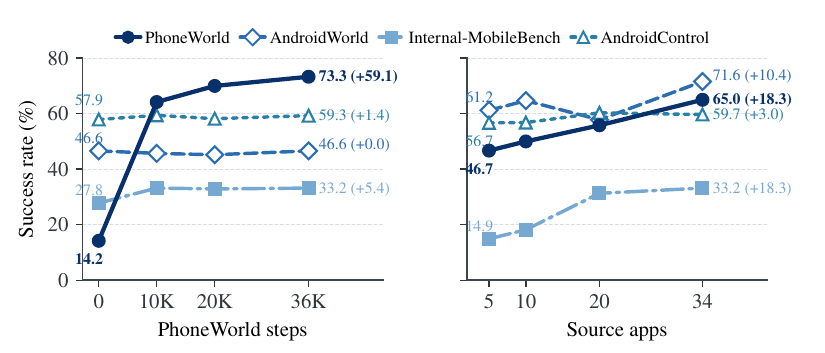}
\caption{\textbf{Scaling PhoneWorld data and app coverage.}
\textbf{Left}: performance as the number of PhoneWorld training steps increases.
\textbf{Right}: performance as the number of source apps increases under a fixed 10K-step
PhoneWorld budget.}
\label{fig:scaling_results}
\end{figure*}

\vspace{-5pt}
\paragraph{Scaling data volume and environment coverage.}
\label{sec:scaling_amount}
Figure~\ref{fig:scaling_results} separates rollout volume from environment diversity. Adding
0/10K/20K/36,193 PhoneWorld steps raises PhoneWorld success monotonically from 14.2 to
64.2/70.0/73.3. Over the same sweep, Internal-MobileBench rises from 27.8 to 33.2 and
AndroidControl from 57.9 to 59.3, while AndroidWorld remains near 46\%. More verified experience
therefore improves in-domain competence, yields modest offline-transfer gains, and does not
measurably erode AndroidWorld performance. With the PhoneWorld budget fixed at 10K steps,
expanding coverage from 5 to 34 apps raises PhoneWorld from 46.7 to 65.0,
Internal-MobileBench from 14.9 to 33.2, and AndroidWorld from 61.2 to 71.6, while
AndroidControl remains broadly stable at 56.7--59.7. Because data volume is fixed, these gains
directly demonstrate the value of environment diversity. Together, the two sweeps show that more
rollouts primarily strengthen in-domain performance, whereas broader app coverage produces the
clearest cross-benchmark gains.

\subsection{Mock-app RL strengthens real-app RL}
\label{sec:rl_case_study}
\begin{table}[t]
\vspace{-15pt}
\centering
\setlength{\belowcaptionskip}{4pt}
\small
\setlength{\tabcolsep}{4.5pt}
\caption{Real-phone and AndroidWorld evaluation. The upper block reports general-agent
references. The lower block compares different training recipes. Single-App, Cross-App, and Mini-App each contain 50 human-judged
tasks. Avg.\ is the unweighted mean of the four reported scores.}
\label{tab:rl_case_study}
\begin{tabular}{l c c c c c}
\toprule
\midrule
\textbf{Model / recipe} & \textbf{Single-App} & \textbf{Cross-App} &
\textbf{Mini-App} & \textbf{AndroidWorld} & \textbf{Avg.} \\
\midrule
Gemini 3.1 Pro & 50.0 & 48.0 & 58.0 & 80.2 & 59.1 \\
GPT-5.4 & 50.0 & 32.0 & 40.0 & 70.7 & 48.2 \\
Claude Opus 4.7 & 38.0 & 16.0 & 28.0 & 56.0 & 34.5 \\
Seed 2.0 Pro & 44.0 & 30.0 & 60.0 & 71.5 & 51.4 \\
\midrule
SFT & 34.0 & 22.0 & 54.0 & 60.3 & 42.6 \\
Real-app RL & 54.0 & 20.0 & 48.0 & 77.2 & 49.8 \\
Real-app + PhoneWorld mock-app RL & 62.0 & 18.0 & 56.0 & 83.2 & 54.8 \\
\midrule
\bottomrule
\end{tabular}
\end{table}

The general-agent block provides reference points on the same evaluations: Gemini 3.1 Pro has
the highest average at 59.1\%, while Seed 2.0 Pro is strongest on Mini-App tasks at
60.0\%. These rows are contextual references rather than controlled comparisons with the
three training recipes.

Under the same 50-step RL budget, allocating half of the rollouts to PhoneWorld improves the real-phone average from
40.67\% to 45.33\% and AndroidWorld from 77.2\% to 83.2\% over real-app RL alone. Gains are
concentrated in single-app and mini-app workflows, while cross-app performance remains weak.
Because the total budget is unchanged, the result shows that verifier-backed mock-app RL
strengthens, rather than merely enlarges, real-app RL training. Appendix
Figure~\ref{fig:rl_showcase} provides qualitative examples of the resulting improvements in
constraint following and cross-app information transfer.

\subsection{Four-app Real--Mock fidelity case study}
\label{sec:fidelity_analysis}
\begin{wraptable}{r}{0.45\columnwidth}
\vspace{-26pt}
\centering
\setlength{\belowcaptionskip}{4pt}
\footnotesize
\setlength{\tabcolsep}{0.4pt}
\caption{\textbf{Real--Mock app fidelity}}
\label{tab:mock-real-fidelity-pilot}
\begin{tabular*}{\linewidth}{@{\hspace{3pt}\extracolsep{\fill}}lrrrrrr@{\hspace{3pt}}}
\toprule
\midrule
\textbf{App} &
\textbf{RPC} &
\textbf{NPC} &
\textbf{FPC} &
\textbf{DER} &
\textbf{TER} &
\textbf{VF} \\
\midrule
App1 & 99.2 & 96.6 & 78.4 & 40.1 & 54.1 & 63.7 \\
App2 & 97.7 & 94.9 & 76.7 & 77.1 & 82.5 & 70.0 \\
App3 & 96.3 & 92.7 & 80.4 & 78.4 & 95.6 & 67.5 \\
App4 & 97.1 & 82.9 & 51.0 & 57.7 & 70.5 & 60.3 \\
\midrule
\bottomrule
\end{tabular*}
\end{wraptable}
We assess reconstruction fidelity on Douban, Xiaohongshu, Bilibili, and Meituan Waimai using
963 real trajectories, 6,442 valid page visits, and 94 page types. RPC, NPC, and FPC measure
rendered, navigable, and functional page coverage; DER and TER measure recurring and
high-frequency transition recall; and VF measures visual similarity. All metrics are weighted by
observed usage; Appendix~\ref{app:fidelity_protocol} provides full definitions.
Aggregated across the four apps, rendered, navigable, and functional coverage reaches 97.7\%,
92.4\%, and 73.0\%, respectively. High-frequency routes are preserved better than the broader
transition graph (73.7\% TER versus 59.7\% DER), with 65.0\% pooled visual fidelity. PhoneWorld
thus provides a \emph{task-oriented reconstruction} that preserves common pages and workflows as
executable interactions while selectively abstracting deeper service-backed functionality and
visual detail.

\section{Conclusion}
PhoneWorld converts real-app screenshots and trajectories into dynamic, resettable, and
verifiable Android environments by preserving task-relevant pages, navigation, visible semantics,
and state effects rather than cloning full services. The resulting suite serves simultaneously as
an online evaluation environment, a verified data-production environment, and an online-RL
training environment through a common reset and verification interface. Across fidelity, SFT, and
RL studies, PhoneWorld provides training signal beyond the mock apps and is most effective as a
scalable complement to real-app experience, suggesting that progress in phone-use agents depends
on scaling controllable environments as well as data.

\subsection*{AI use statement}
Generative AI tools were used as research instruments in the PhoneWorld pipeline to assist screenshot taxonomy construction, page-level specification drafting, Android app implementation, and iterative self-review. All generated applications and tasks were checked through compilation, emulator testing, programmatic verification, and final human acceptance testing as described in Section~\ref{sec:method}. Generative AI was also used for limited language editing and conversion of the manuscript into the official ICLR \LaTeX{} template. The authors reviewed all AI-assisted work and take responsibility for the final text, claims, code, data, and artifacts.

\subsection*{Ethics statement}
PhoneWorld environments are research mock-ups rather than production apps. The planned research release uses distinct package names, synthetic seed data, offline emulator-only execution, and ``Mock'' launcher labels where appropriate. Gated APK access is intended for non-commercial academic research, and app or brand references identify interaction patterns rather than affiliation or endorsement.

\subsection*{Reproducibility statement}
Section~\ref{sec:method} describes environment construction, task synthesis, reset logic, and
verification, with worked examples in Appendix~\ref{app:method_details}.
Appendix~\ref{app:experimental_details} summarizes the evaluation benchmarks and metrics, while
Appendix~\ref{app:fidelity_protocol} defines the fidelity audit. Release plans and access links
for the code, benchmark tasks, runner, documentation, and gated mock-APK dataset are provided
in the Availability section below.

\section*{Availability}
We released the PhoneWorld code, benchmark tasks, runner, and documentation at \phoneworldrepo. The mocked apks are constructed by the first author using personal research resource and will be distributed through a gated Hugging Face dataset at \phoneworldapks{} for non-commercial academic research. Each user must accept the dataset terms independently rather than receiving redistributed APK mirrors. The gated APK release uses distinct package names, synthetic data, offline emulator-only execution, and ``Mock'' launcher labels where appropriate to make clear that the apps are research mock environments rather than real consumer apps.

\bibliography{iclr2027_conference}
\bibliographystyle{iclr2027_conference}

\newpage
\appendix
\section{Additional Construction Details}
\label{app:method_details}
This appendix supplements Section~\ref{sec:method} with a worked reconstruction example,
generated page specifications, and additional task-verifier details.

\subsection{Worked Reconstruction Example}
\label{app:worked_example}
\begin{figure*}[h]
\centering
\includegraphics[width=\linewidth]{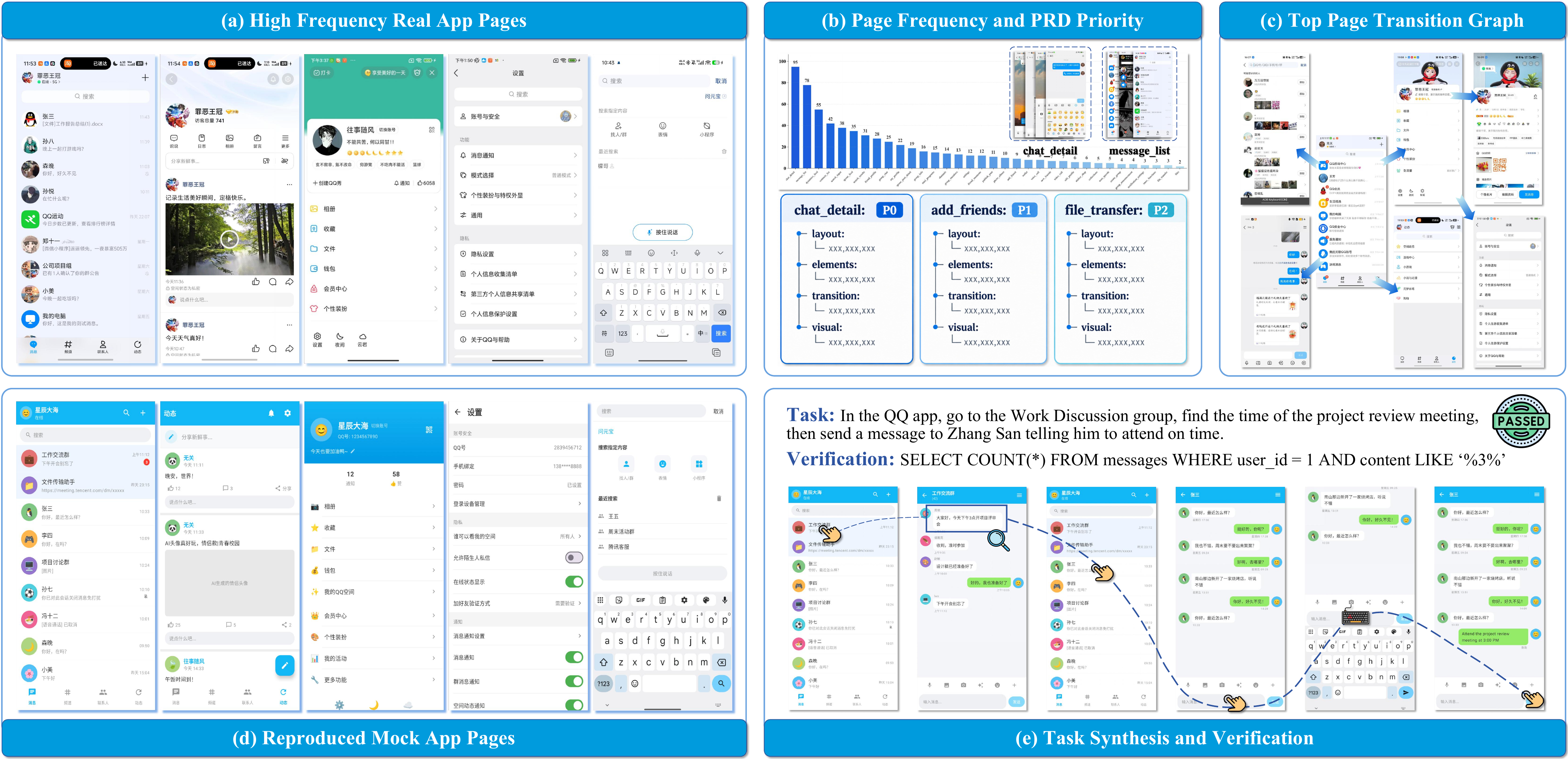}
\caption{\textbf{Worked example of constructing a QQ-like PhoneWorld environment.}
\textbf{(a) High-frequency real-app pages}: representative screenshots of the most visited page types identified from deduplicated app screenshots.
\textbf{(b) Page frequency and build priorities}: trajectory statistics determine page priorities (P0/P1/P2), and each page receives a structured PRD.
\textbf{(c) Transition graph}: dominant navigation flows are extracted from sequential page visits in the trajectories.
\textbf{(d) Generated mock pages}: corresponding screens are implemented from the PRD and reusable component library.
\textbf{(e) Task and verifier}: a representative task requiring information seeking and messaging, checked automatically with a SQLite query.}
\label{fig:qq_case}
\end{figure*}

Figure~\ref{fig:qq_case} grounds the complete construction pipeline in a QQ-like environment.
Starting from representative screenshots and exploratory trajectories, PhoneWorld identifies
behaviorally meaningful page types, aggregates page visitation and transition statistics, and
assigns P0/P1/P2 construction priorities. The resulting skeleton guides page-level PRDs and mock
page implementation; the same executable content and mutable state then ground a representative
task and its SQLite verifier. The example illustrates how observed usage determines both what is
implemented and what can subsequently be executed and verified.

\raggedbottom
\subsection{Generated App Specification Examples}
\label{app:prd_examples}
The following two examples are condensed from the automatically generated Douban PRD. We select
the high-frequency home feed and movie detail pages because together they illustrate both a
navigation-heavy top-level surface and a stateful, information-dense entity page. The examples
retain the complete four-part structure supplied to the construction agent, while condensing
repetitive low-level prose for readability.

\begin{prdcodeblock}{prdgreen}{douban/home\_feed.prd}{P0}{419 visits \(\cdot\) 3 reference views}
\begin{minipage}[t]{0.485\linewidth}
\prdsectiontitle{01}{layout}
\prdcodekv{header}{drawer, central search, audiobook shortcut, and inbox with unread badge}
\prdcodekv{navigation}{Following/Recommended tabs, dismissible banner, five fixed bottom tabs,
and floating compose action}
\prdcodekv{content}{vertical mixed-card feed for group posts, user updates, ratings, and discussions}

\prdcodedivider
\prdsectiontitle{02}{interaction\_and\_state}
\prdcodekv{actions}{search, messages, tabs, cards, identities, join/follow, expand, social actions,
and compose}
\prdcodekv{effects}{join/follow persists membership; dismiss removes the banner; publish inserts
a user-visible feed item}
\end{minipage}
\hfill
\begin{minipage}[t]{0.485\linewidth}
\prdsectiontitle{03}{transitions}
\prdcodekv{search}{\texttt{search\_input}}
\prdcodekv{inbox}{\texttt{messages}}
\prdcodekv{feed cards}{\texttt{group\_post}, \texttt{group\_detail}, \texttt{user\_profile}}
\prdcodekv{bottom tabs}{\texttt{home\_media}, \texttt{home\_groups}, \texttt{home\_profile}}
\prdcodekv{compose}{\texttt{publish}}

\prdcodedivider
\prdsectiontitle{04}{visual}
\prdcodekv{palette}{\texttt{\#FFFFFF}, \texttt{\#2D9144}, \texttt{\#42BD56},
\texttt{\#333333}, \texttt{\#999999}}
\prdcodekv{type}{titles 16--18\,sp; body 14--15\,sp; metadata 12\,sp}
\prdcodekv{geometry}{16\,dp side gutters; 36\,dp avatars; 28\,dp bottom-navigation icons}
\end{minipage}
\end{prdcodeblock}

\begin{prdcodeblock}{prdorange}{douban/movie\_detail.prd}{P0}{285 visits \(\cdot\) 3 reference views}
\begin{minipage}[t]{0.485\linewidth}
\prdsectiontitle{01}{layout}
\prdcodekv{hero}{back/overflow controls, poster, localized/original titles, year, genres, country,
duration, and ranking badges}
\prdcodekv{state panel}{Want/Watched controls, aggregate rating, histogram, and rating counts}
\prdcodekv{detail stream}{playback or ticket sources, synopsis, cast/crew, rankings, and
overview/review/discussion tabs}

\prdcodedivider
\prdsectiontitle{02}{interaction\_and\_state}
\prdcodekv{actions}{back, share, ranking, status controls, rating panel, sources, synopsis,
cast cards, filters, and reviews}
\prdcodekv{effects}{Want/Watched persists in the media library; rating persists both stars and
the required short review}
\end{minipage}
\hfill
\begin{minipage}[t]{0.485\linewidth}
\prdsectiontitle{03}{transitions}
\prdcodekv{rate}{\texttt{rating\_popup}}
\prdcodekv{rating panel}{\texttt{comments\_list}}
\prdcodekv{cast card}{\texttt{person\_detail}}
\prdcodekv{rank/share}{\texttt{ranking\_list}, \texttt{share\_popup}}
\prdcodekv{back}{originating search, ranking, or media-list context}

\prdcodedivider
\prdsectiontitle{04}{visual}
\prdcodekv{palette}{poster-derived hero; white/light-gray foreground; warm orange rating accents}
\prdcodekv{type}{title 20--22\,sp; body 14--16\,sp; metadata 11--12\,sp}
\prdcodekv{geometry}{12--16\,dp section rhythm; poster approximately \(100\times140\)\,dp}
\end{minipage}
\end{prdcodeblock}

\clearpage
\flushbottom
\subsection{Verifier-Coupled Task Synthesis}
\label{app:verifier_details}
PhoneWorld synthesizes tasks jointly from three construction artifacts: read-only content
determines which entities and answers exist, page PRDs determine which interaction paths are
available, and the mutable-state schema determines which outcomes can be verified. The generator
produces a natural-language goal, difficulty level, step limit, reset configuration, and
verification rule, ensuring that every requested entity exists and every required operation is
supported by the app.

\begin{figure}[t]
\centering
\begin{findingbox}{Example of Synthesized Task and Verification Rule}
{\ttfamily\scriptsize
\{\\
\hspace*{0.9em}\textbf{"app"}:~"mqq",\\
\hspace*{0.9em}\textbf{"difficulty"}:~"easy",\\
\hspace*{0.9em}\textbf{"max\_steps"}:~10,\\
\hspace*{0.9em}\textbf{"goal"}:~"In QQ, search for \textbackslash{}"Project Group\textbackslash{}" and tap Favorite.",\\
\hspace*{0.9em}\textbf{"verification"}:~\{\\
\hspace*{1.8em}"type":~"sqlite",\\
\hspace*{1.8em}"database\_path":~"/data/data/com.phoneuse.mqq/databases/qq.db",\\
\hspace*{1.8em}"table":~"user\_collections",\\
\hspace*{1.8em}"rules":~[\\
\hspace*{2.7em}\{ "field":~"target\_type", "operator":~"==", "value":~"group" \},\\
\hspace*{2.7em}\{ "field":~"target\_id", "operator":~"==", "value":~4 \}\\
\hspace*{1.8em}]\\
\hspace*{0.9em}\}\\
\}
}
\end{findingbox}
\caption{Example of a synthesized verifiable task.
Each task is represented as a structured JSON object. The verification field specifies a database-level rule that enables automatic task-completion checking without manual annotation.}
\label{fig:prompt_template}
\vspace{-12pt}
\end{figure}

PhoneWorld uses two deterministic verifier types. Information-seeking tasks check whether the
final response contains values grounded in the read-only content; state-changing tasks query
SQLite for records satisfying task-specific predicates. Cross-app tasks are synthesized from
entities shared by two apps and require retrieving information in one environment before acting
in another. Because every task starts from a known state and admits an automatic outcome check,
the same task interface supports benchmark evaluation and verifier-filtered trajectory
collection.

\paragraph{From verifiers to RL rewards.}
Each verifier also defines a binary terminal reward. Answer-based tasks score the final response,
while state-changing tasks score the final database state. Resettable initialization and
programmatic rewards make PhoneWorld directly usable for repeatable online RL without manual or
model-based judging.

\section{Additional Experimental Details}
\label{app:experimental_details}
This appendix summarizes the PhoneWorld environment suite
(Table~\ref{tab:phoneworld_suite}) and the evaluation settings and metrics used in Section~4.

\begin{table}[htbp]
\centering
\setlength{\belowcaptionskip}{4pt}
\caption{\textbf{Summary of the PhoneWorld environment suite.}}
\label{tab:phoneworld_suite}
\begin{tabular}{l l}
\toprule
\midrule
\textbf{Component} & \textbf{Scale or capability} \\
\midrule
Environment coverage & 34 apps across 16 domains \\
Audited benchmark & 120 tasks: 44 easy + 41 medium + 35 hard \\
Verification & 49 answer-based + 71 SQLite-based tasks \\
Generated task pool & 7,936 verifiable tasks \\
Verified data collection & 3,354 trajectories / 36,193 interaction steps \\
\midrule
\bottomrule
\end{tabular}
\end{table}

\subsection{Benchmarks and Metrics}
Table~\ref{tab:eval_benchmarks} summarizes the four evaluations. PhoneWorld and
AndroidWorld~\citep{rawles2025androidworld} are the primary online benchmarks;
Internal-MobileBench and AndroidControl~\citep{li2024effects} are offline transfer diagnostics.

\begin{table}[t]
\centering
\setlength{\belowcaptionskip}{4pt}
\small
\caption{\textbf{Evaluation benchmarks used in this paper.}}
\label{tab:eval_benchmarks}
\begin{tabular}{l l l p{0.38\linewidth}}
\toprule
\midrule
\textbf{Benchmark} & \textbf{Setting} & \textbf{Metric} & \textbf{Role} \\
\midrule
PhoneWorld & Online mock apps & Task SR & In-domain environment evaluation \\
AndroidWorld & Online real apps & Task SR & Primary real-app transfer \\
Internal-MobileBench & Offline real-device data & Step SR & Practical phone-use transfer \\
AndroidControl & Offline Android data & Step SR & Control-policy transfer \\
\midrule
\bottomrule
\end{tabular}
\end{table}

PhoneWorld contains 44 easy, 41 medium, and 35 hard tasks; the evaluated AndroidWorld split
contains 61 easy, 36 medium, and 19 hard task types. Online success requires full programmatic
credit; partial scores, execution errors, and failed checks count as unsuccessful. The
\emph{Steps} columns in Table~\ref{tab:main_result} report the mean number of dispatched device
actions over solved tasks, excluding screenshot and shell calls. Because this average is
conditioned on success, it measures execution efficiency only within the subset a model solves.

\section{Real--Mock Fidelity Metrics}
\label{app:fidelity_protocol}
We quantify reconstruction fidelity using visit-weighted page coverage, transition recall, and
visual fidelity. All metrics are first computed per app. Aggregate scores pool the corresponding
numerators and denominators across apps rather than averaging per-app percentages.

\paragraph{Page coverage.}
For app \(a\), let \(\mathcal{P}_a\) be its observed real page types, \(n_a(p)\) the number of
real visits to page type \(p\), and \(\ell_a(p)\in\{0,1,2,3\}\) its manually audited mock
implementation level. L0 denotes an absent page; L1 a rendered proxy without a task-equivalent
reachable page; L2 a reachable and useful page with important semantics or actions missing; and L3
a page whose core task-relevant interaction produces the expected state change. For threshold
\(k\), visit-weighted page coverage is
\begin{equation}
\operatorname{PC}_a(k)
=
100\,
\frac{\sum_{p\in\mathcal{P}_a}n_a(p)\,\mathbf{1}[\ell_a(p)\geq k]}
{\sum_{p\in\mathcal{P}_a}n_a(p)}.
\label{eq:fidelity_page_coverage}
\end{equation}
We report RPC \(=\operatorname{PC}_a(1)\), NPC \(=\operatorname{PC}_a(2)\), and FPC
\(=\operatorname{PC}_a(3)\). Thus, RPC measures visible page presence, NPC adds reachability and
task usability, and FPC further requires the core semantic state effect.

\paragraph{Transition recall.}
Let \(\mathcal{E}_a\) be the directed page transitions observed in real trajectories,
\(n_a(e)\) the frequency of transition \(e\), and \(s_a(e)\in\{0,1\}\) indicate whether the
mapped transition is directly supported by the mock according to a static source audit. At a
minimum real-transition count \(c\), frequency-weighted edge recall is
\begin{equation}
\operatorname{ER}_a(c)
=
100\,
\frac{\sum_{e\in\mathcal{E}_a:\,n_a(e)\geq c}n_a(e)s_a(e)}
{\sum_{e\in\mathcal{E}_a:\,n_a(e)\geq c}n_a(e)}.
\label{eq:fidelity_edge_recall}
\end{equation}
DER is \(\operatorname{ER}_a(2)\), which evaluates the broader recurring transition graph, and
TER is \(\operatorname{ER}_a(10)\), which isolates the most frequently traversed navigation
backbone. Weighting by \(n_a(e)\) makes a missed common route count more than a missed rare route.

\paragraph{Visual fidelity.}
For each paired real--mock page type, we assign
\(z_{a,p,d}\in\{0,1,2\}\) on five dimensions \(d\): layout, key elements, affordance, content
semantics, and visual hierarchy. Scores 0, 1, and 2 respectively indicate a substantial mismatch,
a partial match with noticeable differences, and a close task-relevant match. Let
\(\mathcal{M}_a\subseteq\mathcal{P}_a\) denote the paired page types and \(D=5\). Visit-weighted
visual fidelity is
\begin{equation}
\operatorname{VF}_a
=
100\,
\frac{\sum_{p\in\mathcal{M}_a} n_a(p)\sum_{d=1}^{D}z_{a,p,d}}
{2D\sum_{p\in\mathcal{M}_a}n_a(p)}.
\label{eq:fidelity_visual}
\end{equation}

\clearpage
\section{Qualitative Analysis of Real+Mock RL}
\label{app:rl_showcase}
\begin{figure}[h]
\centering
\includegraphics[width=\linewidth]{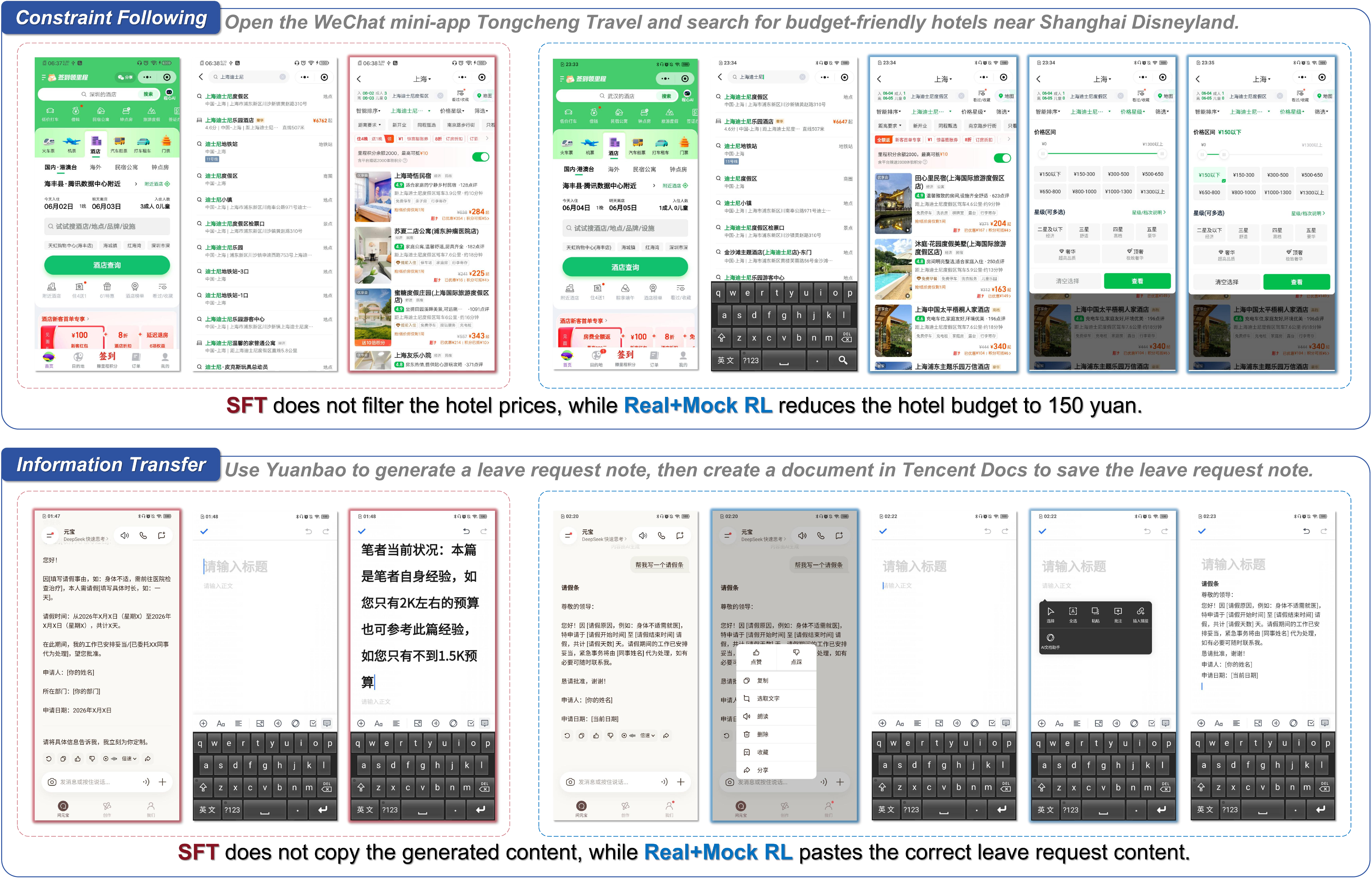}
\caption{\textbf{Qualitative comparison between SFT and Real+Mock RL on real-phone tasks.}
\textbf{Top}: Real+Mock RL applies the requested hotel-price constraint, whereas SFT stops at an
unfiltered result page. \textbf{Bottom}: Real+Mock RL correctly transfers generated text from
Yuanbao to Tencent Docs, whereas SFT inserts stale clipboard content.}
\label{fig:rl_showcase}
\end{figure}

Figure~\ref{fig:rl_showcase} presents two representative real-phone trajectories that illustrate
how Real+Mock RL changes execution behavior beyond the aggregate success rates reported in
Section~\ref{sec:rl_case_study}. The examples compare the SFT policy with the policy trained by
combining real-app and PhoneWorld mock-app rollouts under the same total RL budget.

\paragraph{Constraint following.}
The first task asks the agent to search for budget-friendly hotels near Shanghai Disneyland in
the Tongcheng Travel WeChat mini-app. The SFT policy reaches a plausible hotel-results page but
does not enforce the requested price constraint. The Real+Mock RL policy instead continues to the
filtering interface and sets the maximum hotel price to 150 yuan. This case illustrates improved
retention of task constraints across a multi-step workflow and more complete use of controls that
are not available on the initial search page.

\paragraph{Cross-app information transfer.}
The second task requires the agent to generate a leave-request note with Yuanbao and save it in
Tencent Docs. The SFT policy reaches the document editor but fails to copy the newly generated
note, causing stale clipboard content to be inserted. The Real+Mock RL policy correctly selects
and copies the generated note, switches to Tencent Docs, creates a document, and pastes the
intended content. The improvement therefore involves not only navigation across apps, but also
maintaining and transferring an intermediate information artifact through the clipboard.

Together, these examples suggest that mixed-environment RL does more than encourage broader
exploration: it provides useful supervision for satisfying explicit task constraints and
executing stateful operations required for cross-app information transfer.

\end{document}